\documentclass[conference]{IEEEtran}
\usepackage{multirow}
\IEEEoverridecommandlockouts

\usepackage{cite}
\usepackage{amsmath,amssymb,amsfonts}
\usepackage{algorithmic}
\usepackage{graphicx}
\usepackage{textcomp}
\usepackage{xcolor}
\usepackage{booktabs}
\usepackage{stfloats}
\usepackage{etoolbox}
\usepackage{geometry}

\def\BibTeX{{\rm B\kern-.05em{\sc i\kern-.025em b}\kern-.08em
    T\kern-.1667em\lower.7ex\hbox{E}\kern-.125emX}}

\title{Deployment-friendly Lane-changing Intention Prediction Powered by Brain-inspired Spiking Neural Networks
}

\author{
\IEEEauthorblockN{1\textsuperscript{st} Shuqi Shen}
\IEEEauthorblockA{\textit{The Hong Kong University of} \\
\textit{Science and Technology (Guangzhou)}\\
Guangzhou, China \\
u202141021@xs.ustb.edu.cn}
\and
\IEEEauthorblockN{1\textsuperscript{st} Junjie Yang}
\IEEEauthorblockA{\textit{The Hong Kong University of} \\
\textit{Science and Technology (Guangzhou)}\\
Guangzhou, China \\
jyang512@connect.hkust-gz.edu.cn}
\and
\IEEEauthorblockN{3\textsuperscript{rd} Hui Zhong}
\IEEEauthorblockA{\textit{The Hong Kong University of} \\
\textit{Science and Technology (Guangzhou)}\\
Guangzhou, China \\
hzhong638@connect.hkust-gz.edu.cn}

\and
\IEEEauthorblockN{4\textsuperscript{th} Hongliang Lu*}
\IEEEauthorblockA{\textit{The Hong Kong University of} \\
\textit{Science and Technology (Guangzhou)}\\
Guangzhou, China \\
hlu592@connect.hkust-gz.edu.cn\\
*Corresponding author}
\and
\IEEEauthorblockN{5\textsuperscript{th} Xinhu Zheng*}
\IEEEauthorblockA{\textit{The Hong Kong University of} \\
\textit{Science and Technology (Guangzhou)}\\
Guangzhou, China \\
xinhuzheng@hkust-gz.edu.cn\\
*Corresponding author}
\and
\IEEEauthorblockN{6\textsuperscript{th} Hai Yang}
\IEEEauthorblockA{
\textit{The Hong Kong University of }\\
\textit{Science and Technology}\\
Hongkong, China \\
cehyang@ust.hk\\
}

\thanks{Shuqi Shen and Junjie Yang contribute equally to this paper.}
\thanks{Corresponding authors: Hongliang Lu and Xinhu Zheng.}
}
\begin{document}

\maketitle

\begin{abstract}
Accurate and real-time prediction of surrounding vehicles’ lane-changing intentions is a critical challenge in deploying safe and efficient autonomous driving systems in open-world scenarios. Existing high-performing methods remain hard to deploy due to their high computational cost, long training times, and excessive memory requirements.
Here, we propose an efficient lane-changing intention prediction approach based on brain-inspired Spiking Neural Networks (SNN). By leveraging the event-driven nature of SNN, the proposed approach enables us to encode the vehicle's states in a more efficient manner.
Comparison experiments conducted on HighD and NGSIM datasets demonstrate that our method significantly improves training efficiency and reduces deployment costs while maintaining comparable prediction accuracy. Particularly, compared to the baseline, our approach reduces training time by 75\% and memory usage by 99.9\%. These results validate the efficiency and reliability of our method in lane-changing predictions, highlighting its potential for safe and efficient autonomous driving systems while offering significant advantages in deployment, including reduced training time, lower memory usage, and faster inference.
\end{abstract}

\begin{IEEEkeywords}
Lane-changing intention prediction, Spiking Neural Networks (SNN), autonomous driving, real-time prediction, HighD dataset, NGSIM dataset.
\end{IEEEkeywords}

\section{Introduction}
%在自动驾驶领域，快速准确地预测周围车辆驾驶员的变道意图是提升系统安全性和决策效率的关键。精准的意图预测不仅能有效避免潜在碰撞，还能优化路径规划和交通流管理，实现更智能高效的驾驶体验。
%先说一句自动驾驶对于未来交通或者出行的意义，算是一个套话
In the field of autonomous driving, the rapid and accurate prediction of lane-changing intentions of surrounding vehicles is critical for enhancing system safety and decision-making efficiency. Precise intention prediction can not only effectively prevent potential collisions but also optimize path planning and traffic flow management, resulting in a smarter and more efficient driving experience \cite{mozaffari2020deep}.

Recent advancements in lane-changing prediction research have been remarkable. Unlike earlier approaches that relied on physical models that simulate vehicle dynamics through predefined mathematical equations \cite{kesting2007general, wang2014investigation}, modern research has shifted toward learning-based methods. For instance, back-propagation neural networks effectively predict lane-changing intentions by learning from real-world data \cite{ding2013neural}. Similarly, temporal models such as Long Short-Term Memory (LSTM) networks have gained popularity for their ability to capture temporal dependencies in lane-changing prediction tasks \cite{shokrolah2019trajectory}. The Echo State Network (ESN), a more recent temporal model, further improves prediction performance through its unique dynamic properties \cite{griesbach2021lane}.
Learning-based methods for lane-changing prediction, despite their advancements, face notable limitations. They require extensive training data, involve prolonged training durations, and demand significant hardware resources, posing challenges for real-time deployment on onboard devices. These issues highlight the need for more efficient and lightweight solutions \cite{xie2019data}.
Furthermore, the continuous influx of massive amounts of data amplifies the need for fast algorithm iterations and real-time processing capabilities. As a result, lane-changing prediction algorithms must balance efficient training, rapid deployment, and real-time reliability to meet the demands of practical applications.

% 受大脑启发的方法推动了神经计算领域的重大进步，SNN 就是此类创新的典型例子。 SNN 被认为是第三代神经网络 \cite{ghosh2009third}，通过基于脉冲的事件驱动机制进行操作。这种受生物学启发的方法不仅反映了神经处理的自然稀疏性，而且还展示了高能源效率。这些特性使得 SNN 非常适合资源受限的嵌入式系统，并且能够有效地应对动态交通场景的复杂性，利用其时间动态进行实时决策。
% Brain-inspired methods have spurred significant advancements in neural computation, with SNN standing out as a prime example of such innovation. Recognized as the third-generation neural networks \cite{ghosh2009third}, SNN operate through spike-based event-driven mechanisms. This biologically inspired approach not only mirrors the natural sparsity of neural processing but also demonstrates high energy efficiency. 
% %
% These characteristics render SNN highly suitable for resource-constrained embedded systems and effective in navigating the complexities of dynamic traffic scenarios, leveraging their temporal dynamics for real-time decision-making.
Brain-inspired computational methods have recently made groundbreaking progress. SNN, known as the third-generation neural network \cite{ghosh2009third}, exhibits unique advantages \newgeometry{top=0.75in, bottom=0.8in, left=0.75in, right=0.75in}in intelligent embedded systems due to biological interpretability and deployment efficiency. Unlike traditional Artificial Neural Networks (ANN), SNN is event-driven, relying on a sparse spike timing encoding mechanism to process information. This event-driven feature enhances computational energy efficiency by 12 times, as it only processes information when events occur, reducing unnecessary computation cost 
\cite{kundu2021spike}. Therefore, numerous researchers have shown great interest in SNN.
\cite{neftci2019surrogate} introduced the surrogate gradient method into SNN, significantly improving training speed by over threefold, while also enhancing deployment efficiency. \cite{rueckauer2017conversion} developed an ANN-SNN conversion framework, which maintained model accuracy while increasing training efficiency by 75\%, and also reduced resource consumption. %所以呢，罗列完例子应该总结一下
These researches collectively highlight the significant potential of SNN in achieving both high computational and deployment efficiency and reliable performance, making SNN a promising solution for future real-world practical applications.

In this paper, we propose an efficient lane-changing intention prediction method based on brain-inspired SNN. 
Specifically, our approach first utilizes a linear layer to feature the vehicle’s driving state. Then, we employ SNN to understand and predict the vehicle’s lane-changing intention in real-time, categorizing it into no lane-keeping, turn left, or turn right.
To evaluate our approach, several groups of comparison experiments are carried out.
we utilize two open-source naturalistic driving datasets, HighD \cite{krajewski2018highd} and NGSIM \cite{coifman2017critical}, to assess the performance in terms of efficiency and accuracy. 
The advantages of SNN in resource consumption support the development of highly efficient algorithms and seamless hardware deployment, enabling scalable and practical autonomous driving solutions.
Looking ahead, brain-inspired approaches like SNN stand out as a key advantage for autonomous driving, offering a paradigm shift in mimicking the brain’s adaptive and efficient problem-solving capabilities. 

The remainder of this paper is organized as follows. Section II discusses the related work. Section III introduces the proposed methodology in detail. Section IV presents the experimental results and analyzes the findings. Finally, Section V concludes the paper and highlights potential directions for future research.

\section{Related Work}
% 总起讲这一节回顾了什么
This section reviews previous work on lane-changing prediction and SNN. We first discuss the development of lane-changing prediction methods, comparing kinematic or kinetic models and behavioral models with learning-based approaches, and highlight their limitations. Next, we introduce SNN, emphasizing its advantages and applications, particularly in traffic and autonomous vehicle domains.

\subsection{Lane-Changing Intention Prediction}
% 换道预测是指在交通场景中，预测其他车辆是否会进行换道操作以及何时进行换道的任务。
Lane-changing prediction refers to predicting whether and when other vehicles will execute a lane change in a traffic environment.
% 早期的换道预测方法主要基于运动学或动力学模型，通过数学公式进行建模, 例如
% XX使用极坐标多项式对车辆轨迹进行拟合并做出预测，但由于多项式无法充分捕捉复杂的非线性动态，导致无法完全拟合车辆的真实轨迹。
% 非线性动力学模型和模型预测控制的应用确保了纵向安全和横向操控稳定，但这种方法未考虑交通参与者的动态预测，限制了其在复杂交通环境中的实际应用。
Early lane-changing prediction approaches were primarily based on kinematic or kinetic models, typically using mathematical formulas for modeling. For example, \cite{nelson1989continuous} used polar coordinate polynomials to fit vehicle trajectories and predict their motion, demonstrating an effective mathematical approach for trajectory prediction. Nonlinear dynamic models and model predictive control were utilized to ensure longitudinal safety and lateral stability in lane-changing scenarios, showcasing their potential in maintaining vehicle control under various conditions \cite{liu2018dynamic}.
% 此外, 行为模型在换道预测中也得到了广泛验证, 例如  研究重点模拟了驾驶员的行为决策过程，通过明确选择目标车道并评估换道安全性的方法提高了换道预测的准确性，但在复杂的交通场景下的适用性不足。
Besides, behavioral modeling methods were also widely utilized. For example, \cite{treiber2009modeling} modeled the driver’s decision-making process and improved lane-changing prediction accuracy by explicitly selecting target lanes and evaluating lane-change safety.
% 这些模型严重依赖模型的复杂度, 高复杂度模型虽然实现了高预测准确率但在部署实时性上面临严重不足
These models achieve high prediction accuracy by relying heavily on their complexity, however, the computational demands make them unsuitable for resource-constrained onboard deployment in real-time applications.
% 近年来，基于学习的方法逐渐成为换道预测的主流。例如
In recent years, learning-based methods have gradually become the mainstream in lane-changing prediction.
% xx使用了多层感知机 MLP学习驾驶轨迹并进行换道预测, 实现了较高的准确率, 但是无法预测完整路径的换道预测
% 结合双向 RNN和 LSTM 对时间序列驾驶数据进行学习也能够对高速公路换道意图进行准确的预测, 然而模型的复杂使得其部署效率无法满足实时需求
% XX 使用了回声状态网络（Echo State Networks, ESNs）和主成分分析（PCA）对自然驾驶数据进行学习，实现了对换道行为的高精度预测。然而，这种方法依赖于驾驶员相关数据，在预测周围车辆（SV）的换道意图时存在局限性。
% 针对驾驶数据不足的问题，xxx 提出了一种基于在线迁移学习的换道意图预测方法，在数据量有限的情况下成功实现了目标车辆换道意图的检测，但在实际应用中，尤其是右侧换道行为的预测准确性仍有待提升。
For example, \cite{tomar2010prediction} employed a multilayer perceptron to learn driving trajectories and predict lane changes, demonstrating the potential of neural networks for capturing lane-changing patterns. Similarly, the combination of bidirectional RNN and LSTM proved effective for predicting highway lane-changing intentions using time-series driving data, showcasing the strength of temporal models in handling sequential information \cite{xing2020ensemble}.
\cite{griesbach2020prediction} advanced lane-changing prediction by applying ESN and Principal Component Analysis to naturalistic driving data, leveraging driver-specific data for precision. Additionally, \cite{zhang2021target} introduced an online transfer learning approach that efficiently predicted lane-changing intentions even with limited data, expanding the applicability of learning-based methods in diverse scenarios.
% overall, 这些基于学习的方法无法实现高效的训练和部署与高预测准确率之间的有效平衡
Overall, these learning-based methods fail to achieve both efficient training and deployment and high prediction accuracy at the same time.
Therefore, a deployment-friendly approach for efficient lane-changing intention prediction is needed.

\begin{figure*}[t]
    \centering
    \includegraphics[width=0.85\textwidth]{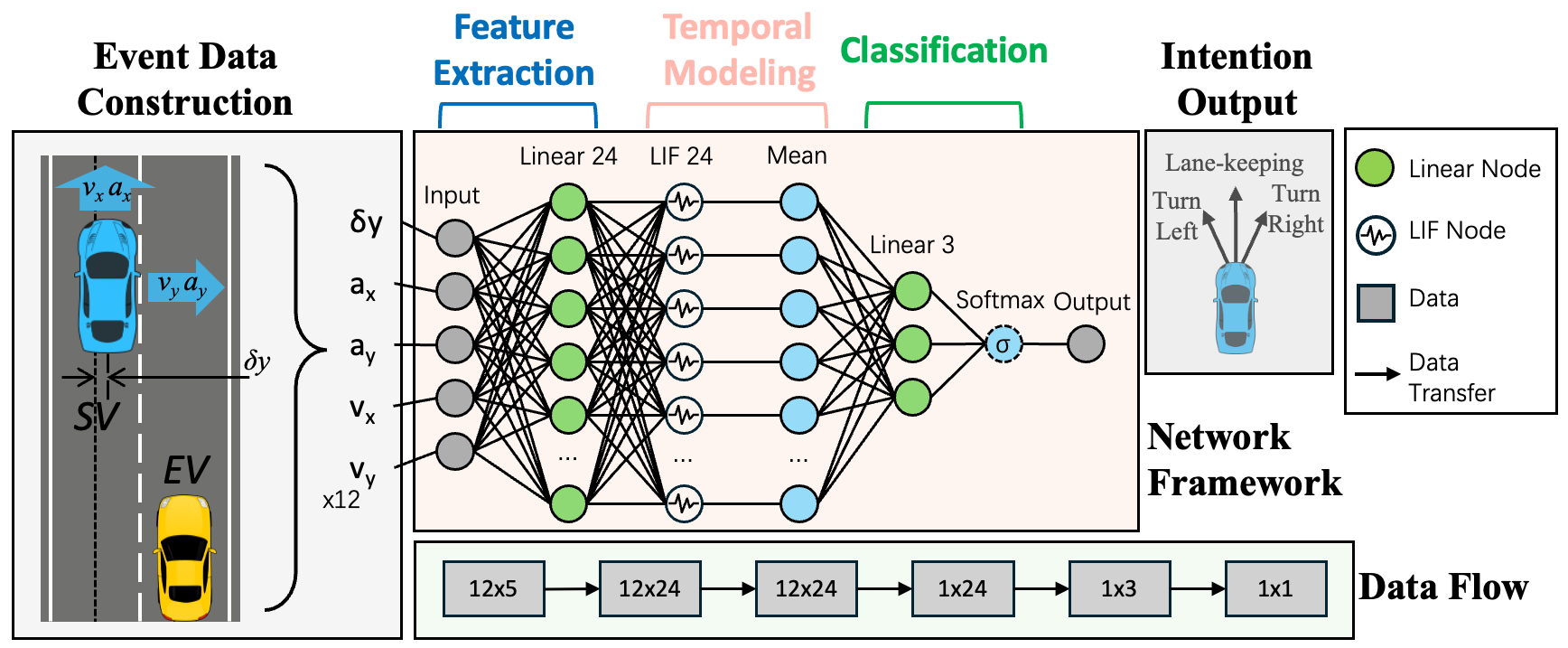}
    \caption{Schematic of the SNN model for lane-changing intention prediction. The model processes a time-series input of vehicle state features through four main components: (1) Event Data Construction to generate the input matrix, (2) Feature Extraction with a Linear layer to expand feature dimensions, (3) Temporal Modeling using a LIF layer to capture temporal dependencies, and (4) Classification where a Linear layer maps the features to lane-change intention categories, followed by a Softmax activation for prediction.}
    \label{framework}
\end{figure*}

\subsection{Spiking Neural Networks}
% 脉冲神经网络（SNN）是一种高效、低功耗且具有生物可解释性的神经网络，特别适合处理时间序列数据和资源受限的实时应用。
SNN are efficient, low-power, and biologically interpretable neural networks that are particularly suited for processing time-series data and real-time applications in resource-constrained environments.
% 近年来 SNN 在很多领域得到了广泛的应用
% Fang等人在时间序列分类任务中使用脉冲神经网络（SNN），通过设计稀疏时空脉冲编码方案和训练算法，在低功耗的前提下实现了与深度神经网络相当的分类性能。
% Kim等人在动态视觉传感（DVS）任务中使用脉冲神经网络（SNN），实现了深层SNN在多个基准数据集上的最先进优化性能。
% []等人通过使用SNN完成视觉识别任务，在SNN导向的数据集上表现出更优的性能，并显著降低了能耗。
% []在复杂视觉识别任务中提出了一种新的 ANN-SNN 转换机制，在确保识别准确率的同时显著提升了训练效率。
In recent years, SNN have been widely applied in many fields. For example, \cite{fang2020multivariate} utilized SNN for time series classification and designed a sparse spatiotemporal spike encoding scheme and training algorithm, achieving classification performance comparable to deep neural networks under low-power conditions. \cite{kim2021optimizing} utilized SNN for dynamic vision sensing (DVS) optimization tasks, achieving optimal performance for deep SNN across multiple benchmark datasets. \cite{deng2020rethinking} used SNN for visual recognition tasks and demonstrated superior performance on SNN-specific datasets with a significant reduction in energy consumption. \cite{sengupta2019going} proposed a new ANN-SNN conversion mechanism for complex visual recognition tasks that ensured recognition accuracy and significantly improved training efficiency.
% 脉冲神经网络在交通领域的应用
In the field of transportation, SNN have also garnered significant attention.
% 在其他交通参与者的分类任务中，SNN与事件摄像头的结合实现了低功耗和低延迟的高效分类性能。
% []在自动驾驶车辆的车道保持任务中应用SNN，并结合STDP实现了快速学习，展现出更高的定位精度，同时显著降低了计算能耗。
In the classification tasks of other traffic participants, the combination of SNN and event-based cameras demonstrated efficient classification capabilities with low power consumption and low latency \cite{viale2021carsnn}.
\cite{lopez2021spiking} proposed novel SNN for AV radar signal processing, which achieved efficient processing performance and maintained low power consumption in simulated driving scenarios.
\cite{bing2020indirect} applied SNN to lane-keeping tasks to achieve fast learning, which improved positioning accuracy and significantly reduced computational energy consumption.
% 总而言之 xx
In summary, SNN has shown great promise in transportation applications, meeting the critical demands of autonomous driving for fast, energy-efficient deployment.

\section{Methodology}
% 总起
This section introduces the proposed approach for efficient lane-changing intention prediction using SNN, including the network architecture and the training process.

\subsection{Proposed SNN Architecture}

To capture the temporal dynamics in lane-changing processes, we propose a model based on Spiking Neural Networks (SNN), as illustrated in Figure \ref{framework}. The input to the model consists of a time-series matrix representing vehicle state features with dimensions $(12, 5)$. Each row corresponds to a time step, and the five feature dimensions include lateral distance to the lane center($\delta_y$), longitudinal velocity($v_x$), longitudinal acceleration($a_x$), lateral velocity($v_y$), and lateral acceleration($a_y$). These features provide critical spatial and dynamic information required for understanding lane-changing behavior.
During dataset construction, lane-changing initiation is defined as the moment when the vehicle starts deviating from its lane. It is assumed that drivers form clear lane-changing intentions 3 s before this moment \cite{jokhio2023analysis}.

The model architecture integrates three main components to enable efficient and accurate prediction of lane-changing intentions. First, the feature extraction module, implemented as a Linear layer, expands the input feature dimensions, enhancing the representation power of the data. Next, the temporal modeling module, implemented as a Leaky Integrate-and-Fire (LIF) layer, captures the dynamic dependencies in the input data by simulating biological neuron behaviors. Finally, the classification module aggregates the extracted temporal features and maps them into a probabilistic output representing three categories of lane-changing intentions. This architecture ensures that the temporal patterns in the input data are preserved and efficiently processed, enabling accurate predictions while maintaining computational efficiency.

\subsection{Model Training}
The training process involves detailed data transformations and computations within each component of the model. The input to the model is a time-series matrix $\mathbf{X} \in \mathbb{R}^{[12, 5]}$ of vehicle state features, which first undergoes a feature expansion step through a linear layer. This layer transforms the input dimensions from $(12, 5)$ to a $(12, 24)$ matrix $\mathbf{I} \in \mathbb{R}^{[12, 24]}$ by applying a linear transformation:
\begin{equation}
\mathbf{I} = \mathbf{X}\mathbf{W} + \mathbf{b}
\end{equation}
where $\mathbf{W}$ and $\mathbf{b}$ are the weight matrix and bias vector of the linear layer. This expansion improves the expressive power of the features.

The expanded feature matrix is then processed by the LIF layer, which models temporal dependencies using a neuron state update mechanism. Each neuron processes a specific feature dimension independently, with its state updated iteratively across time steps:
\begin{equation}
u_j[t] = \beta \cdot u_j[t-1] + I[t,j]
\end{equation}
where $\beta \in (0,1)$ is the decay coefficient that controls the retention of historical states. When the neuron’s state exceeds a predefined threshold, it emits a spike, resetting its state:
\begin{equation}
    s_j[t] =
    \begin{cases}
    1, & u_j[t] \geq V_{\text{th}} \\
    0, & \text{otherwise}
    \end{cases}
\end{equation}

The output of the LIF layer is a spike matrix $\mathbf{S} \in {0,1}^{[12, 24]}$. To extract global temporal features, the spike matrix is averaged along the time dimension:
\begin{equation}
\bar{S}j = \frac{1}{12} \sum{t=1}^{12} S_j[t], \quad \forall j \in {1, \dots, 24}
\end{equation}
This produces a $(1, 24)$ feature vector, which is then passed through a linear layer that maps the features to a 3-dimensional output corresponding to the three lane-changing intention categories. Finally, a Softmax activation function is applied to convert the outputs into a probability distribution:
\begin{equation}
P_i = \frac{\exp(z_i)}{\sum_{k=1}^{3} \exp(z_k)}, \quad i \in {0, 1, 2}
\end{equation}
where $z_i$ is the $i_{th}$ output, and $P_i$ represents the predicted probability of the $i_{th}$ lane-changing intention.

The model is trained using supervised learning, with the goal of minimizing the difference between the predicted probability distribution and the true labels. Negative log-likelihood loss (NLLLoss) is used as the loss function:
\begin{equation}
\mathcal{L} = - \frac{1}{N} \sum_{i=1}^N \log P(a_t^{(i)} | s_t^{(i)})
\end{equation}
where  $a_t^{(i)}$  represents the true lane-changing intention at time step  $t $ for the  $i_{th}$  sample, and $ s_t^{(i)}$  is the corresponding input state. $ P(a_t | s_t)$  denotes the predicted probability of the true class, and  N  is the batch size, set to 128 in this study.

Through iterative optimization, the model learns to effectively capture the dynamic patterns in vehicle state features, leading to improved performance in lane-changing intention prediction tasks.

\section{Experimental Result}
This section presents the experimental results to evaluate the effectiveness and efficiency of the proposed SNN-based lane-changing intention prediction model in comparison with the ESN method and LSTM method. We train our model on HighD and NGSIM datasets, and the evaluation is conducted from three perspectives: training performance, two sample cases, and overall evaluation. All experiments, including model training and testing, are conducted on an Apple M2 CPU. First, we analyze the model’s training process by comparing the time required per epoch and loss, demonstrating the superior efficiency of our approach. Second, we evaluate the model’s practical application in specific scenarios, showing that our predictions align closely with the ground truth lane-changing moments, outperforming baseline methods in dynamic traffic environments. Finally, an overall evaluation is presented, highlighting the performance of our method compared to ESN and LSTM. 

\subsection{Model Training}

\begin{figure}[h]
    \centering
    \includegraphics[width=0.45\textwidth]{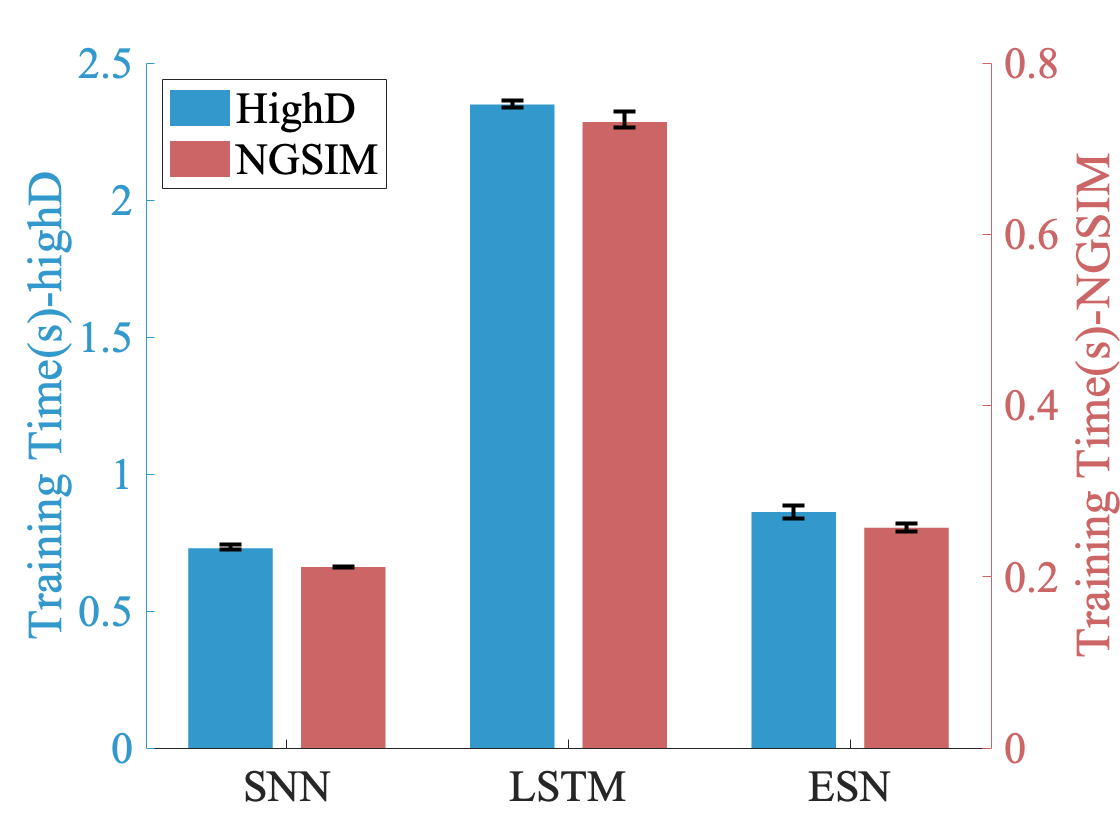}
    \caption{Comparison of average training time per epoch for our approach, LSTM, and ESN on the HighD and NGSIM datasets.}
    \label{time}
\end{figure}
\begin{figure}[h]
    \centering
    \includegraphics[width=0.5\textwidth]{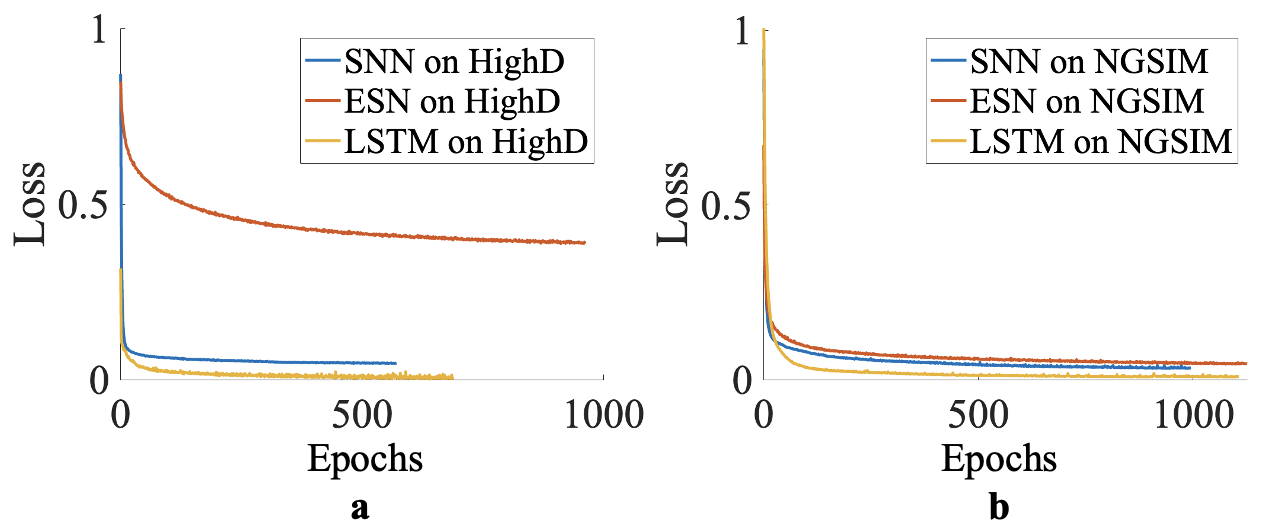}
    \caption{Loss curves for SNN, ESN, and LSTM on the HighD (a) and NGSIM (b) datasets}
    \label{loss}
\end{figure}
% 在本节中，我们展示了我们方法的训练效果，包括训练时间和 loss 曲线的表现，并与 ESN 和 LSTM 方法进行了对比。
In this section, we present the training performance of our approach, ESN, and LSTM methods, including the training time and the loss curve. We train the models of the three approaches on the full HighD and NGSIM datasets.

% 首先，我们展示了训练时间的表现。我们在 HighD 和 NGSIM 两个数据集上分别进行了 500 个 epoch 的训练，结果如图\ref{time}所示。图中蓝色坐标轴表示 HighD 数据集上的训练时间，红色坐标轴表示 NGSIM 数据集上的训练时间。柱状图展示了 500 个 epoch 中每轮训练时间的平均值，误差条（error bar）则展示了训练时间的最大最小值。在 HighD 数据集中，我们的方法每轮平均训练时间为 0.72 秒，LSTM 方法为 2.38 秒，ESN 方法为 0.84 秒；在 NGSIM 数据集中，我们的方法每轮平均训练时间为 0.016 秒，LSTM 方法为 0.048 秒，ESN 方法为 0.018 秒。实验结果表明，我们的方法在训练效率上显著优于 LSTM 方法，同时略优于 ESN 方法，尤其是在 NGSIM 数据集上表现尤为突出。
First, we present the training time performance. We train the three models on the HighD and NGSIM datasets for 500 epochs and show the training time results, as shown in Figure \ref{time}. The blue axis represents the training time on the HighD dataset, while the red axis represents the training time on the NGSIM dataset. The bar chart illustrates the average training time per epoch across 500 epochs, with the blue bars representing the HighD dataset and the red bars representing the NGSIM dataset. The error bars indicate the maximum and minimum training times. On the HighD dataset, our approach achieves an average training time of 0.72 s per epoch, compared to 2.38 s for LSTM and 0.84 s for ESN. On the NGSIM dataset, our approach achieves 0.211 s per epoch, compared to 0.731 s for LSTM and 0.257 s for ESN. These results demonstrate that our approach significantly outperforms LSTM in training efficiency and slightly outperforms ESN, particularly on the NGSIM dataset.

% 接着，我们展示了模型的 loss 曲线。训练过程中，当 loss 值在 50 轮内不再下降时，我们停止训练。如图\ref{loss}所示，横坐标表示训练轮次，纵坐标表示 loss 值。蓝色曲线代表我们的方法，红色和黄色曲线分别代表 ESN 和 LSTM 方法。图\ref{loss}a 展示了三种方法在 HighD 数据集上的 loss 曲线：我们的方法在 522 轮后收敛于 0.045，ESN 方法在 912 轮后收敛于 0.387，LSTM 方法在 639 轮后收敛于 0.0015。图\ref{loss}b 展示了三种方法在 NGSIM 数据集上的 loss 曲线：我们的方法在 1492 轮后收敛于 0.050，ESN 方法在 852 轮后收敛于 0.225，LSTM 方法在 1489 轮后收敛于 0.0。实验结果表明，我们的方法在 HighD 数据集上收敛速度显著快于 ESN 和 LSTM 方法，而在 NGSIM 数据集上，尽管收敛轮次较长，但最终 loss 值与 LSTM 方法接近，且显著优于 ESN 方法。
Next, we present the loss curves of the models. During training, we stop when the loss value does not decrease within 50 epochs. As shown in Figure \ref{loss}, the horizontal axis represents the number of training epochs, and the vertical axis represents the loss value. The blue curve represents our method, the red curve represents the ESN method, and the yellow curve represents the LSTM method. Figure \ref{loss}a shows the loss curves for the three methods on the HighD dataset. Our method converges at 0.045 after 522 epochs, while the ESN method converges at 0.387 after 912 epochs, and the LSTM method converges at 0.0015 after 639 epochs. Figure \ref{loss}b shows the loss curves on the NGSIM dataset. Our method converges at 0.032 after 945 epochs, the ESN method converges at 0.044 after 1076 epochs, and the LSTM method converges at 0.0 after 1090 epochs. The results indicate that our method achieves significantly faster convergence on the HighD dataset compared to ESN and LSTM. On the NGSIM dataset, although our method requires more epochs to converge than ESN, the final loss value is comparable to LSTM and significantly better than ESN.

% 总而言之，我们的方法在训练效率和模型性能上均表现出色。与 LSTM 方法相比，我们的方法在训练时间上具有显著优势，同时 loss 收敛性能与 LSTM 方法相当；与 ESN 方法相比，我们的方法在 loss 收敛性能和训练时间上明显更优。这证明了我们的方法在兼顾高效训练和优异性能方面的潜力。
In summary, our method performs exceptionally well in training efficiency. Compared to the LSTM method, our method shows a significant advantage in training time while achieving comparable loss convergence performance. Compared to the ESN method, our method demonstrates superior performance in both loss convergence and training time. These results highlight the potential of our method to achieve efficient training and excellent performance simultaneously.

\begin{figure*}[t]
    \centering
    \includegraphics[width=0.9\textwidth]{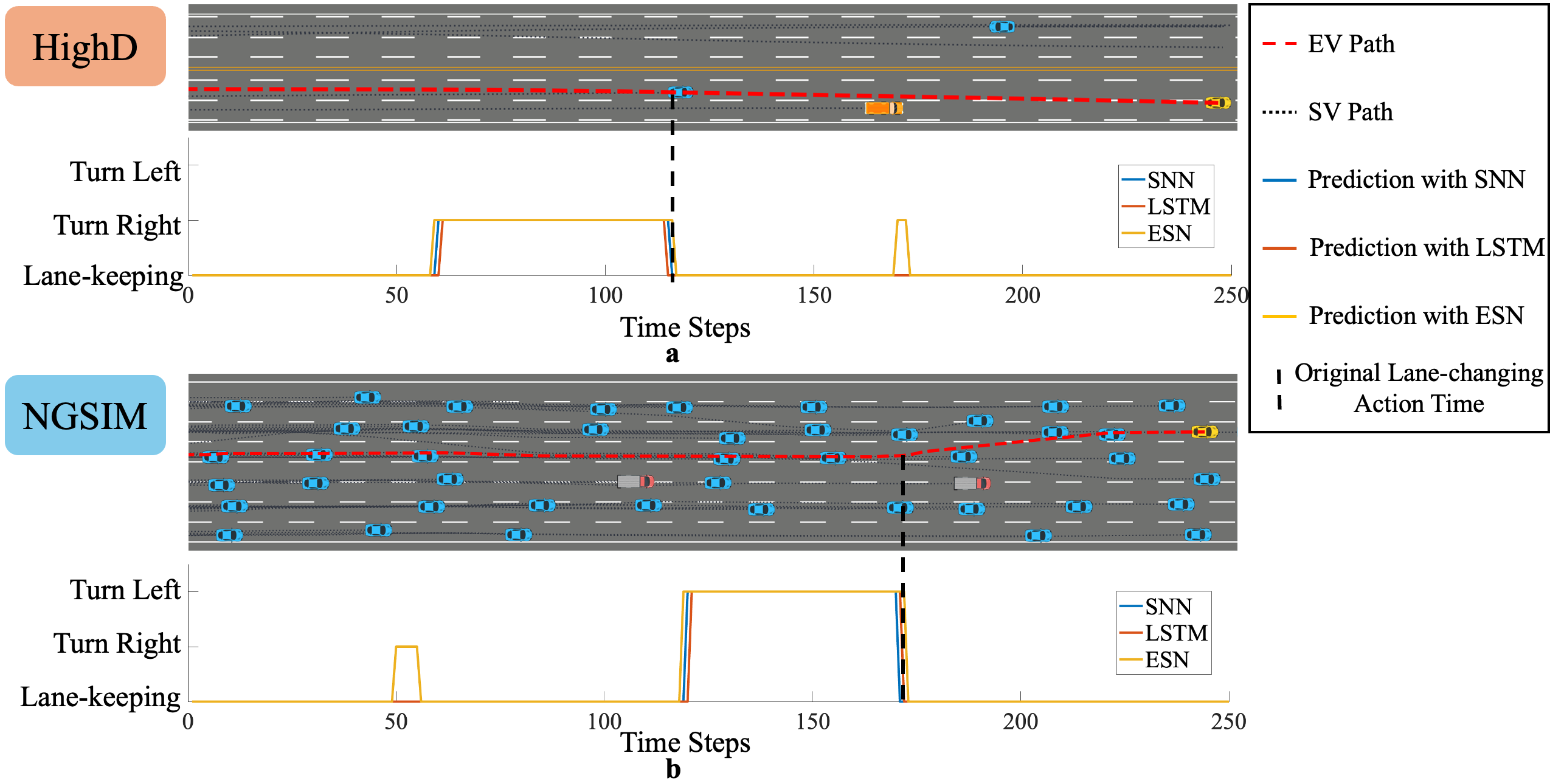}
    \caption{Comparison of lane-changing intention predictions for SNN, LSTM, and ESN models in the HighD (a) and NGSIM (b) datasets. The blue, red, and yellow curves represent predictions from the SNN, LSTM, and ESN models, respectively. Our method predicts lane-changing accurately, with no false predictions.}
    \label{case}
\end{figure*}

\subsection{Two Sample Cases}

This section illustrates the deployment performance of our SNN model compared to the ESN and LSTM models in the HighD and NGSIM scenarios through two sample cases. The analysis includes scenario visualizations and lane-changing intention prediction curves. In Figure \ref{case}, the selected vehicles are highlighted in yellow, with their trajectories represented by red dashed lines. Other vehicles are marked with different colors, and their trajectories are represented by black dotted lines. The lane-changing intention prediction curves are shown, with black dashed lines indicating the lane-changing moments in the original data. The blue curve represents the predictions of our method, while the red and yellow curves represent the predictions from the LSTM and ESN methods, respectively.

Figure \ref{case}a shows a scenario from the HighD dataset, where a vehicle turns right at time-step 115. Experimental results show that the proposed method predicts the lane-changing intention at time-step 60, the LSTM method at time-step 61, and the ESN method at time-step 59. However, the ESN method makes a false prediction at time-step 170.
Figure \ref{case}b shows a scenario from the NGSIM dataset, where a vehicle turns left at time-step 170. Experimental results show that the proposed method predicts the lane-changing intention at time-step 120, the LSTM method at time-step 121, and the ESN method at time-step 119. However, the ESN method produces a significant false prediction at time-step 50.

In summary, the comparative experiments demonstrate that the proposed method performs exceptionally well in lane-changing intention prediction tasks. Our method produces no false predictions and maintains stable performance across different datasets and scenarios. These results confirm the effectiveness and reliability of the proposed method in practical applications.

\subsection{Overall Evaluation}

\begin{table*}[b]
\centering
\caption{Comparison of Models on HighD and NGSIM Datasets}
\label{tab:comparison}

\begin{tabular}{@{}lcccccc@{}}
\toprule
\textbf{Model} & \textbf{Dataset} & \textbf{Parameters} & \textbf{Memory Usage} & \textbf{Training Time} & \textbf{Accuracy} \\ \midrule
\textbf{SNN}   & HighD            & \textbf{219}                 & \textbf{0.01 MB}              & \textbf{375 s}                 & 0.9828                                 \\
\textbf{ESN}   & HighD            & 1,503               & 0.04 MB              & 766 s                 & 0.8876                                        \\
\textbf{LSTM}  & HighD            & 269,059             & 80.81 MB             & 1520 s                 & \textbf{0.9983}                                        \\ \midrule
\textbf{Model} & \textbf{Dataset} & \textbf{Parameters} & \textbf{Memory Usage} & \textbf{Training Time} & \textbf{Accuracy}  \\ \midrule
\textbf{SNN}   & NGSIM            & \textbf{219}                 & \textbf{0.01 MB}              & \textbf{199 s }               & 0.9426                                   \\
\textbf{ESN}   & NGSIM            & 1,503               & 0.04 MB              & 276 s                & 0.9043                                       \\
\textbf{LSTM}  & NGSIM            & 269,059             & 80.81 MB             & 796 s                &\textbf{ 0.9809 }                                  \\ \bottomrule
\end{tabular}%

\end{table*}
% 本节我们展示了我们的方法的整体表现, 并与 esn 和 LSTM 方法进行对比, 表 1 显示了这三种方法在参数量, 内存占用, 整体训练时间, 准确率上的表现

% 从表中可知, 我们的方法的参数量仅为 xx, 内存占用为xx, 在HighD 数据集上 xxx s 完成收敛, 准确率达到了 xxx, 而在 ngsim 数据集上 xxx s 完成收敛, 准确率达到了 xxx
% ESN方法的参数量仅为 xx, 内存占用为xx, 在HighD 数据集上 xxx s 完成收敛, 准确率达到了 xxx, 而在 ngsim 数据集上 xxx s 完成收敛, 准确率达到了 xxx
% LSTM方法的参数量仅为 xx, 内存占用为xx, 在HighD 数据集上 xxx s 完成收敛, 准确率达到了 xxx, 而在 ngsim 数据集上 xxx s 完成收敛, 准确率达到了 xxx

% 结果表明, 我们的方法在参数量,内存占用和训练时间上远高于另外两种方法, 体现出极高的训练效率和部署优势, 我们的方法在准确率上也远高于ESN 方法, 虽然略低于 LSTM 方法,xxx

\begin{figure}[!h]
    \centering
    \includegraphics[width=0.49\textwidth]{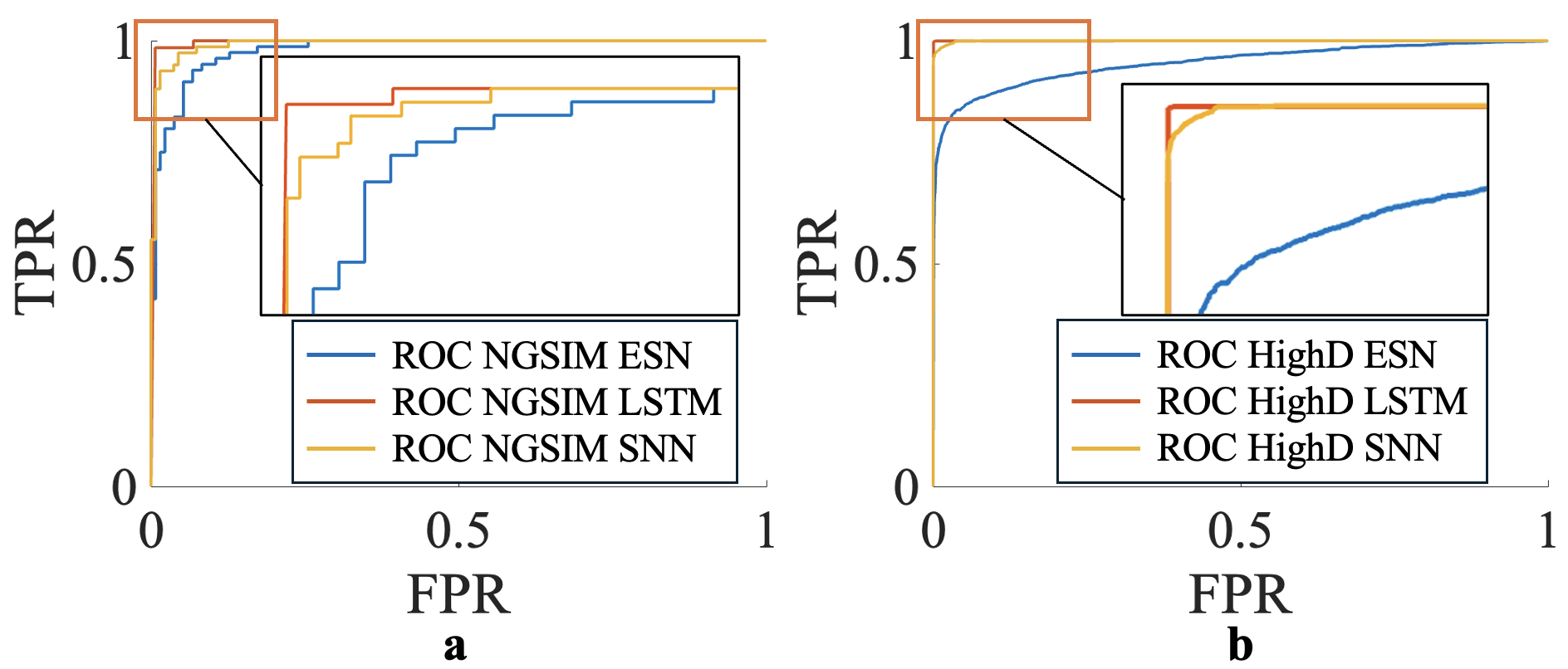}
    \caption{ROC curves for SNN, LSTM, and ESN on the NGSIM (a) and HighD (b) datasets. The blue, red, and yellow curves represent the SNN, LSTM, and ESN methods, respectively.}
    \label{roc}
\end{figure}

This section presents the overall performance of our method compared to ESN and LSTM, including Receiver Operating Characteristic (ROC) curves and a summary table of the overall results. The ROC curves in Figure \ref{roc} illustrate the prediction accuracy \cite{mandrekar2010receiver}. Table \ref{tab:comparison} summarizes the results in terms of parameters, memory usage, training time, and accuracy.

First, Figure \ref{roc} presents the accuracy results of the three methods. The horizontal axis represents the False Positive Rate (FPR), which is the proportion of negative samples incorrectly predicted as positive. The FPR is calculated as:
\begin{equation}
\text{FPR} = \frac{\text{FP}}{\text{FP} + \text{TN}}
\end{equation}
where $\text{FP}$ (False Positive) is the number of negative samples incorrectly predicted as positive, and $\text{TN}$ (True Negative) is the number of negative samples correctly predicted as negative. The vertical axis represents the True Positive Rate (TPR), which is the proportion of positive samples correctly predicted as positive. The TPR is calculated as:
\begin{equation}
\text{TPR} = \frac{\text{TP}}{\text{TP} + \text{FN}}
\end{equation}
where $\text{TP}$ (True Positive) is the number of positive samples correctly predicted as positive, and $\text{FN}$ (False Negative) is the number of positive samples incorrectly predicted as negative.

The curves in Figure \ref{roc} are the ROC curves, which reflect the model’s performance under different classification thresholds. The ROC curve plots the TPR against the FPR at various classification thresholds, and the area under the curve (AUC) is used to quantify the overall performance. A ROC curve closer to the top-left corner, with a higher AUC value, reflects better classification accuracy \cite{moses1993combining}.
Figure \ref{roc}a shows the ROC curves of the three methods on the NGSIM dataset. The AUC value of our method is 0.9925, the AUC value of the LSTM method is 0.9955, and the AUC value of the ESN method is the lowest at 0.9793. Figure \ref{roc}b shows the results on the HighD dataset. The AUC value of our method is 0.9994, the LSTM method achieves 0.9997, and the ESN method achieves 0.9503.
These results demonstrate that our method performs excellently in prediction accuracy, achieving results comparable to the LSTM method while significantly outperforming the ESN method. Moreover, our method significantly outperforms the LSTM approach in terms of training efficiency, which underscores the advantage of our approach in effectively balancing both high accuracy and efficiency.

Then, as shown in Table \ref{tab:comparison}, our method has only 219 parameters, requires 0.01 MB of memory, converges in 375 s on the HighD dataset with an accuracy of 0.9828, and converges in 199 s on the NGSIM dataset with an accuracy of 0.9426. 
The ESN method has 1,503 parameters, requires 0.04 MB of memory, converges in 766 s on the HighD dataset with an accuracy of 0.8876, and converges in 276 s on the NGSIM dataset with an accuracy of 0.9043. 
The LSTM method, with 269,059 parameters and 80.81 MB of memory usage, achieves convergence in 1520 s on the HighD dataset with an accuracy of 0.9983, and in 796 s on the NGSIM dataset with an accuracy of 0.9809. These results demonstrate that our method significantly outperforms the ESN method in terms of parameters, memory usage, and training time while achieving high accuracy. Compared to the LSTM method, our approach offers substantial advantages in efficiency and resource utilization, with only a slight trade-off in accuracy.

In summary, the proposed method shows a compelling balance between accuracy and deployment efficiency. It not only delivers prediction performance comparable to LSTM but also demonstrates significant advantages in terms of resource consumption, training speed, and ease of deployment. Our approach is particularly well-suited for real-time applications and resource-limited environments, offering substantial benefits in both accuracy and deployment efficiency.

\section{Conclusion}
% 本文提出了一种基于脉冲神经网络（SNN）的高效换道意图预测方法，旨在解决自动驾驶系统中对周围车辆换道行为的实时准确预测问题。通过结合SNN的时间动态特性和特征融合能力，我们的方法在保证高预测准确率的同时，显著提升了训练和部署效率。

% 实验结果表明，与LSTM和ESN方法相比，本文方法在训练效率上分别提高了约67%和66%，同时在预测准确率上与LSTM方法相当，显著优于ESN方法。在场景验证中，我们的方法能够更准确地捕捉换道时机，其预测结果与实际换道时刻的偏差最小，展现了在不同数据集和复杂场景下的稳定性能。

% 未来，我们将进一步优化模型架构，探索更高效的训练算法，并尝试将本文方法集成到实际自动驾驶系统中，验证其在真实道路环境中的性能表现。本文的研究为自动驾驶系统的安全决策提供了可靠的技术支持。
This paper proposes an efficient lane-changing intention prediction approach based on SNN, addressing the critical challenge of real-time and accurate prediction of surrounding vehicles' lane-changing behavior in autonomous driving systems. By leveraging the temporal dynamics and feature fusion capabilities of SNN, our method achieves high prediction accuracy while significantly improving training and deployment efficiency. The proposed approach reduces training time by 75\% and memory usage by 99.9\% compared to baseline methods, demonstrating its potential for practical deployment in resource-constrained environments.
Experimental results show that, compared to LSTM and ESN methods, the proposed method achieves prediction accuracy comparable to LSTM and significantly outperforms ESN. In scenario validation, the proposed method accurately captures lane-changing timing, with the smallest deviation from actual lane-changing moments.

Future work will focus on optimizing the model architecture to improve both accuracy and efficiency. We will explore more scenarios to enhance the model’s adaptability. Efforts will also be made to improve prediction performance in diverse conditions. Finally, we aim to test on real-world vehicles to validate the practical effectiveness.

\section*{Acknowledgment}
This study is supported by RGC General Research Fund (GRF) HKUST16205224 and NSFC Grant U24A20252 and 62373315 and the Red Bird MPhil Program at the Hong Kong University of Science and Technology (Guangzhou)

\bibliographystyle{IEEEtran}
\bibliography{main}

\end{document}